\documentclass[11pt]{article}

\usepackage[preprint]{acl}

\usepackage{times}
\usepackage{latexsym}

\usepackage[T1]{fontenc}
\usepackage[utf8]{inputenc}

\usepackage{microtype}

\usepackage{inconsolata}

\usepackage{graphicx}
\usepackage{subfigure}
\usepackage{amsmath}
\usepackage{amssymb}
\usepackage{geometry}
\usepackage{multirow}
\usepackage{colortbl}  

\usepackage{arydshln} 

\definecolor{resqgreen}{HTML}{D5F0D5} 
\title{CoQuant: Joint Weight-Activation Subspace Projection for Mixed-Precision LLMs}

\author{
\textbf{Zhe Ding\textsuperscript{1}},
\textbf{Su Pan\textsuperscript{1}\thanks{Corresponding author.}},
\textbf{Duowei Pan\textsuperscript{2}}
\\
\textsuperscript{1}School of Internet of Things, Nanjing University of Posts and Telecommunications, Nanjing, China \\
\textsuperscript{2}Amazon AGI, Seattle, WA, USA \\
\texttt{\{2024070902, supan\}@njupt.edu.cn}, 
\texttt{dwpan@amazon.com}
}

\begin{document}
\maketitle
\begin{abstract}
Post-training quantization (PTQ) has become an important technique for reducing the inference cost of Large Language Models (LLMs). While recent mixed-precision methods improve ultra-low bit quantization by preserving critical subspaces in high precision, they typically construct these subspaces relying solely on activation statistics. This ignores the fundamental nature of linear operations, where the output perturbation is jointly driven by both activation and weight quantization noise. In this paper, we propose CoQuant, a joint weight-activation subspace projection method. By theoretically modeling the expected output error, CoQuant formulates a closed-form weighted PCA solution that balances activation and weight covariances to select the optimal high-precision subspace. Extensive experiments on Llama-3.2 and Qwen2.5 models show that CoQuant consistently outperforms strong PTQ baselines in both WikiText perplexity and zero-shot common-sense reasoning accuracy. These results demonstrate that joint weight-activation subspace modeling provides a principled and effective direction for low-bit LLM quantization. The source code is available at \url{https://github.com/Zachary5895/CoQuant}.
\end{abstract}

\section{Introduction}

In recent years, Large Language Models (LLMs) have achieved remarkable progress in tasks such as natural language processing, complex reasoning, and code generation \citep{achiam2023gpt,roziere2023code,shao2024deepseekmath}. These models have demonstrated strong generalization across diverse tasks, highlighting their broad applicability. However, as model sizes scale to hundreds of billions of parameters, the inference process imposes prohibitive demands on GPU memory footprint, memory bandwidth, and computational resources \citep{sheng2023flexgen,kwon2023efficient,patel2024splitwise,yang2025xquant}. Furthermore, the growing demand for long-context processing and high-throughput serving exacerbates these hardware bottlenecks, driving up deployment costs \citep{alizadeh2024llm}. Therefore, achieving efficient inference without compromising model accuracy has emerged as a paramount challenge.

To alleviate these inference costs, post-training quantization (PTQ) has emerged as a highly compelling solution, as it obviates the need for expensive retraining while ensuring low deployment overhead and strong adaptability. Early PTQ methods, such as GPTQ \citep{frantar2022gptq} and AWQ \citep{lin2024awq}, primarily focused on weight-only quantization, significantly reducing model storage and memory access overhead by compressing weights to 4-bit or even lower. Furthermore, achieving end-to-end inference acceleration necessitates the simultaneous quantization of activations. However, due to the presence of extreme outliers in the activation distributions of LLMs, direct low-bit uniform quantization often incurs severe accuracy degradation. To suppress these outliers, a series of methods based on orthogonal rotation matrices (e.g., Hadamard transforms) have recently emerged \citep{chee2023quip,ashkboos2024quarot,liu2024spinquant,hu2025ostquant}. By applying computationally equivalent random rotations end-to-end, these methods effectively smooth out activation outliers, unlocking the potential for ultra-low-bit uniform quantization.

Despite significant advances in rotation-based uniform quantization, a notable performance gap persists relative to full-precision baselines at ultra-low bit-widths. Consequently, researchers have turned to mixed-precision quantization (MPQ), aiming to preserve a small fraction of critical elements in high precision (e.g., 8-bit) while compressing the remainder into low precision (e.g., 4-bit). For example, QUIK \citep{ashkboos2024quik} proposed retaining specific outlier channels in activations at high precision. Building upon this, ResQ \citep{pmlr-v267-saxena25b} further modeled the quantization problem from a subspace perspective, systematically reducing quantization error in theory by constructing low-rank orthogonal subspaces and projecting crucial information into the high-precision subspace.

\begin{figure}[t]
  \centering
  \subfigure[Activation-only subspace partition based on activation covariance.]{
    \includegraphics[width=0.8\linewidth]{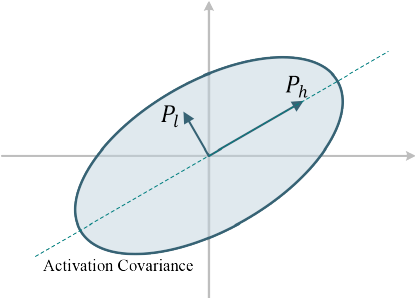}
    \label{fig:fig1a}
  }
  \subfigure[Joint subspace partition based on both activation covariance and weight covariance, where the high-precision direction is adapted to a joint optimum.]{
    \includegraphics[width=0.8\linewidth]{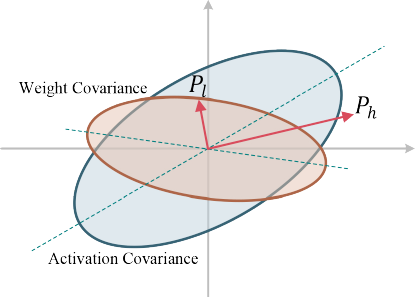}
    \label{fig:fig1b}
  }
  \caption{Subspace selection under different statistical criteria.}
  
\end{figure}

Despite these advances, an important limitation remains. Existing subspace-based MPQ methods for LLMs, as exemplified by ResQ \citep{pmlr-v267-saxena25b}, construct the high-precision subspace primarily from activation statistics. As illustrated in Figure~\ref{fig:fig1a}, their theoretical analysis is based on an upper bound of the activation quantization error, and the high-precision components are selected along the principal directions of the activation covariance matrix. In practice, however, the output error of a linear layer is jointly influenced by perturbations in both activations and weights. Therefore, determining the high-precision subspace solely from activation structure cannot fully capture the nature of joint quantization error.

\begin{figure*}[t]
  \centering
  \includegraphics[width=0.8\textwidth]{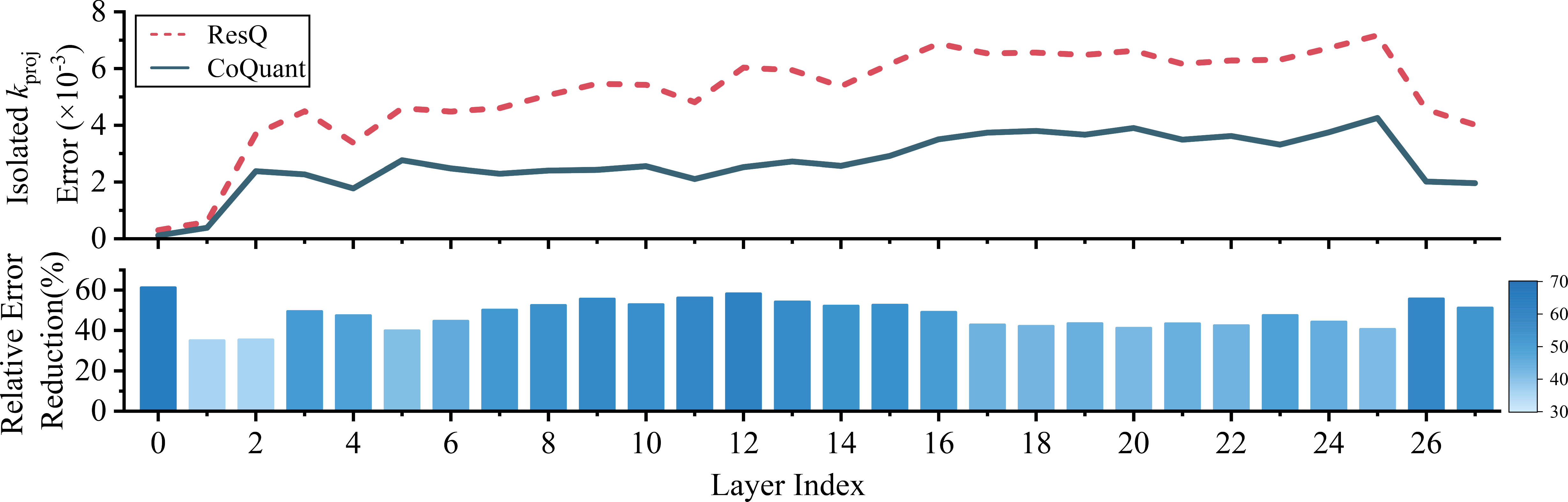}
  \caption{Isolated layer-wise quantization error analysis on the $k_{proj}$ module. To isolate the effect of subspace selection, we fix the rest of the model in full precision and apply mixed-precision quantization exclusively to the weights and input activations of the target $k_{proj}$ layer at each specific index. \textbf{Top:} The absolute output perturbation error comparison between ResQ (relying solely on activation covariance) and our CoQuant (joint weight-activation covariance). \textbf{Bottom:} The relative error reduction achieved by CoQuant. By jointly modeling both statistical dimensions, CoQuant consistently decreases the quantization error by roughly $30\% \sim 60\%$ across all transformer layers.}
  \label{fig:fig2}
\end{figure*}

Motivated by this limitation, we propose \textbf{CoQuant, a joint weight-activation subspace projection method for mixed-precision quantization in LLMs.} Unlike prior subspace-based mixed-precision methods that rely only on activation statistics, CoQuant constructs the high-precision subspace by jointly modeling the quantization effects of both activations and weights, as illustrated in Figure~\ref{fig:fig1b}. Starting from the output error of linear layers, we derive a unified objective for joint weight-activation quantization, which leads to a closed-form subspace solution based on a weighted combination of activation and weight covariance matrices. To illustrate this advantage empirically, Figure~\ref{fig:fig2} provides an isolated layer-wise error analysis on the $k_{proj}$ module, demonstrating that our joint modeling significantly and consistently reduces the actual quantization error across all layers compared to the activation-only baseline (ResQ). In this way, CoQuant provides a more principled and comprehensive foundation for MPQ in LLMs.

To validate the effectiveness of CoQuant, we conduct extensive experiments on multiple LLM families and evaluation benchmarks, covering both language modeling and zero-shot commonsense reasoning tasks. Experimental results show that CoQuant achieves the best overall accuracy-efficiency trade-off, yielding lower perplexity and stronger zero-shot performance across a wide range of model scales. These results demonstrate that jointly modeling weight and activation statistics provides a more effective subspace design for MPQ in LLMs.

\section{Related Work}

In post-training quantization (PTQ) for Large Language Models (LLMs), handling outliers remains the primary bottleneck for achieving high-accuracy low-bit representations. Early works primarily unfolded within the classic PTQ framework. For instance, GPTQ \cite{frantar2022gptq} utilizes approximate second-order information to achieve highly accurate weight quantization, while AWQ \citep{lin2024awq} further introduces the activation-aware weight protection concept to mitigate quantization errors in salient channels. As research extended from weight-only quantization to more general joint weight-activation quantization, systematic outliers in activations emerged as the primary bottleneck. LLM.int8() \citep{llmint8} highlighted the existence of sparse yet systemic outlier features in LLMs and preserved a small number of these anomalous dimensions in high-precision computation through mixed-precision decomposition. Concurrently, SmoothQuant \citep{smoothquant} mitigates the stretching of the low-bit quantization range caused by activation outliers by migrating the quantization difficulty from activations to weights via linear scaling. Furthermore, researchers began leveraging the numerical invariance of orthogonal transformations to improve quantizability by reshaping the representation space rather than explicitly isolating outliers. QuIP \citep{chee2023quip} improves weight-only quantization performance by addressing the incoherence of the weight matrix and its proxy Hessian via randomized orthogonal matrix preprocessing. QuaRot \cite{ashkboos2024quarot} incorporates random Hadamard rotations into the model architecture to suppress outliers in hidden states without altering the full-precision output, achieving end-to-end 4-bit quantization for weights, activations, and KV caches. Building on this, SpinQuant \citep{liu2024spinquant} generalizes random rotations to learnable rotations, directly optimizing the rotation matrices to minimize quantization error. OSTQuant \citep{hu2025ostquant} combines orthogonal transformations with scaling transformations to jointly optimize weight and activation distributions from the perspective of overall quantization space utilization. Overall, PTQ has shifted from error compensation and channel protection toward distribution restructuring through equivalent transformations, improving low-bit quantization stability.

Another relevant line of research is mixed-precision quantization. In traditional CNNs/DNNs, the core philosophy of mixed-precision is to adaptively allocate different bit-widths based on the varying sensitivity of different layers or tensors to quantization errors, rather than applying a uniform precision across the entire network \citep{rakka2024review}. Compared to fixed-precision, mixed-precision better accommodates the differences in redundancy and distribution characteristics across layers, thereby achieving a superior accuracy-efficiency trade-off. Representative works in this direction include DQ \citep{uhlich2019mixed}, which simultaneously learns quantization step sizes and dynamic ranges in a differentiable manner, further inferring the bit-widths for weights and activations across layers. HAQ \citep{wang2019haq} utilizes reinforcement learning combined with hardware feedback to automatically search for layer-wise weight and activation bit-widths. In the context of LLMs, the definition of mixed-precision transcends traditional layer-wise bit allocation; it increasingly manifests as structural designs that retain a small fraction of highly sensitive structures at high precision while compressing the remainder to low precision. LLM.int8() \citep{llmint8} employs mixed-precision decomposition to keep a few outlier features in 16-bit. QUIK \citep{ashkboos2024quik} performs joint weight-activation quantization, compressing most weights and activations to 4-bit while maintaining only a few outlier weights and activations at higher precision. ResQ \citep{pmlr-v267-saxena25b} moves beyond simple outlier channel heuristics by identifying high-variance low-rank subspaces via principal component analysis (PCA), preserving this subspace in high precision and the complementary subspace in low precision, and further suppresses outliers through random rotations within the subspaces. Building upon this paradigm, our work advances subspace-level mixed-precision design by unifying joint weight-activation error modeling to derive an optimal weighted PCA objective for orthogonal subspace decomposition.

\section{Methodology}

\subsection{Background}
\label{subsec:background}
\paragraph{Standard Uniform Quantization.}
Quantization maps a high-precision tensor to a low-precision numerical representation, thereby reducing memory footprint and computation cost. For a full-precision tensor $\mathbf{X} \in \mathbb{R}^{m \times n}$, the simulated $N$-bit uniform quantization operator can be written as
\begin{equation}
    \mathcal{Q}_N(\mathbf{X}) = \operatorname{RoundClip}\left( \frac{\mathbf{X} - z_{\mathbf{X}}}{s_{\mathbf{X}}} \right) \cdot s_{\mathbf{X}} + z_{\mathbf{X}},
\end{equation}
where $\mathcal{Q}_N(\mathbf{X})$ denotes the dequantized tensor after $N$-bit quantization, $s_{\mathbf{X}} \in \mathbb{R}_{+}$ is the quantization scale, and $z_{\mathbf{X}} \in \mathbb{R}$ is the zero-point. The operator $\operatorname{RoundClip}(\cdot)$ rounds each element to the nearest integer and clips it into the valid integer range determined by $N$. In the common symmetric quantization setting, the zero-point is fixed as $z_{\mathbf{X}}=0$, and the scale is typically given by $s_{\mathbf{X}}=\max(|\mathbf{X}|)/(2^{N-1}-1)$; asymmetric quantization can be defined similarly by choosing the scale and zero-point according to the dynamic range of $\mathbf{X}$.

However, uniform quantization is sensitive to outliers. A few extreme values can substantially enlarge the quantization range, causing most normal values to be mapped to only a small number of quantization levels and thereby introducing large quantization error.

\paragraph{Orthogonal Equivalence and Outlier Suppression.}
Consider a standard linear layer
\begin{equation}
    \mathbf{Y} = \mathbf{X}\mathbf{W},
\end{equation}
where $\mathbf{X} \in \mathbb{R}^{n \times d}$ denotes the input activation, $\mathbf{W} \in \mathbb{R}^{d \times m}$ denotes the weight matrix, and $\mathbf{Y} \in \mathbb{R}^{n \times m}$ denotes the output. For any orthogonal matrix $\mathbf{U} \in \mathbb{R}^{d \times d}$ satisfying $\mathbf{U}\mathbf{U}^{\top}=\mathbf{I}$, the linear transformation can be equivalently rewritten as
\begin{equation}
    \mathbf{Y} = \mathbf{X}(\mathbf{U}\mathbf{U}^{\top})\mathbf{W}
    = (\mathbf{X}\mathbf{U})(\mathbf{U}^{\top}\mathbf{W}),
\end{equation}
which leaves the full-precision output unchanged.

This property provides the basis for rotation-based quantization methods. When $\mathbf{U}$ is chosen as a random orthogonal matrix or a Hadamard-based transform, the projected activations and weights tend to become more evenly distributed, and their outlier energy is dispersed across dimensions. As a result, the quantization error is less dominated by a few extreme channels, which improves the robustness of low-bit quantization.

\paragraph{Subspace-wise Mixed-Precision Formulation.}
Building on the orthogonal transform introduced above, MPQ can be formulated by partitioning the transformed feature space into two complementary subspaces \citep{pmlr-v267-saxena25b}. Specifically, let
\begin{equation}
    \mathbf{U} = [\mathbf{U}_l, \mathbf{U}_h] \in \mathbb{R}^{d \times d},
\end{equation}
where $\mathbf{U}_h \in \mathbb{R}^{d \times r}$ spans an $r$-dimensional high-precision subspace and $\mathbf{U}_l \in \mathbb{R}^{d \times (d-r)}$ spans its complementary low-precision subspace. Since $\mathbf{U}$ is orthogonal, we have
\begin{equation}
    \mathbf{U}_l \mathbf{U}_l^\top + \mathbf{U}_h \mathbf{U}_h^\top = \mathbf{I}.
\end{equation}
Accordingly, the activations and weights can be decomposed as
\begin{equation}
    \mathbf{X}_l = \mathbf{X}\mathbf{U}_l,\quad
    \mathbf{X}_h = \mathbf{X}\mathbf{U}_h,
\end{equation}
\begin{equation}
    \mathbf{W}_l = \mathbf{U}_l^\top \mathbf{W},\quad
    \mathbf{W}_h = \mathbf{U}_h^\top \mathbf{W},
\end{equation}
which yields the following subspace decomposition of the layer output:
\begin{equation}
    \mathbf{Y} = \mathbf{X}_l \mathbf{W}_l + \mathbf{X}_h \mathbf{W}_h.
\end{equation}

A mixed-precision scheme can then be defined by quantizing the two subspaces with different bit-widths. Let $\mathcal{Q}_l(\cdot)$ and $\mathcal{Q}_h(\cdot)$ denote the low-precision and high-precision quantizers, respectively. The quantized output is written as
\begin{equation}
    \hat{\mathbf{Y}} =
    \mathcal{Q}_l(\mathbf{X}_l)\mathcal{Q}_l(\mathbf{W}_l)
    +
    \mathcal{Q}_h(\mathbf{X}_h)\mathcal{Q}_h(\mathbf{W}_h).
\end{equation}
Under this formulation, subspace-wise MPQ reduces to determining an appropriate orthogonal basis and selecting the subspace that should be preserved in higher precision.

\subsection{CoQuant: Joint Weight-Activation Subspace Projection}
\label{sec:coquant}

\begin{figure}[t]
  \centering
  \includegraphics[width=0.8\linewidth]{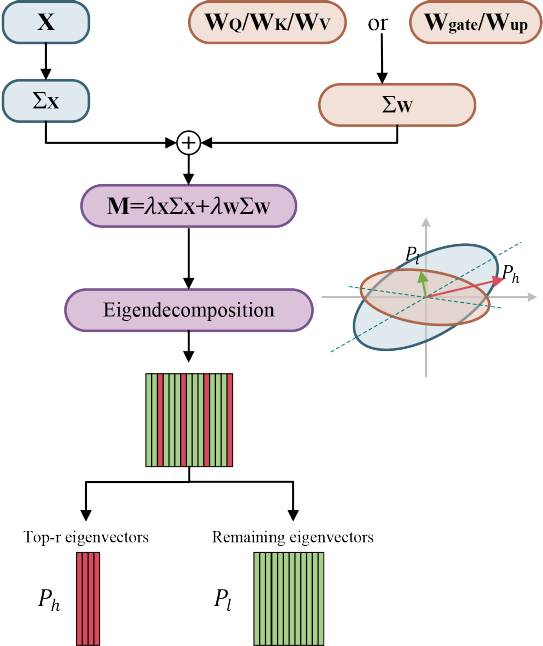}
  \caption{Overview of CoQuant. Given the input activation covariance $\Sigma_{\mathbf{X}}$ and the fused weight covariance $\Sigma_{\mathbf{W}}$ (constructed from shared-input linear layers such as $\mathbf{W}_{\mathbf{Q}}/\mathbf{W}_{\mathbf{K}}/\mathbf{W}_{\mathbf{V}}$ or $\mathbf{W}_{\textbf{gate}}/\mathbf{W}_{\textbf{up}}$), CoQuant forms the mixed covariance matrix $\mathbf{M}=\lambda_{\mathbf{X}}\Sigma_{\mathbf{X}}+\lambda_{\mathbf{W}}\Sigma_{\mathbf{W}}$ and performs eigendecomposition to obtain a joint subspace partition. The top-$r$ eigenvectors define the high-precision subspace $\mathbf{P}_h$, while the remaining eigenvectors form the low-precision subspace $\mathbf{P}_l$.}
  \label{fig:fig3}
\end{figure}

Existing subspace-based methods, such as ResQ \citep{pmlr-v267-saxena25b}, construct the high-precision subspace by minimizing the activation quantization error alone. However, in linear layer operations $\mathbf{Y} = \mathbf{XW}$, the output perturbation is intrinsically coupled with the quantization errors of both activations and weights. Therefore, selecting the subspace solely based on activation principal components is suboptimal. We propose \textbf{CoQuant}, which directly minimizes the expectation of the joint output error.

\paragraph{Joint Error Modeling.} 
Following the subspace partition defined in Section~\ref{subsec:background}, let the quantized components be $\hat{\mathbf{X}}_k = \mathbf{X}_k + \mathbf{E}_{\mathbf{X}_k}$ and $\hat{\mathbf{W}}_k = \mathbf{W}_k + \mathbf{E}_{\mathbf{W}_k}$ for $k \in \{l, h\}$, where $\mathbf{E}$ denotes the respective quantization noise. 
To model the joint error, we adopt the standard additive quantization noise model (AQNM) \cite{widrow2008quantization}. Crucially, because our framework applies invariant random orthogonal rotations within the subspaces, the projected tensors are Gaussianized and outliers are suppressed. This structural property motivates the AQNM approximation, where the quantization noise is modeled as zero-mean and approximately independent of the signal. Furthermore, by treating the quantization errors as small perturbations, the second-order term $\mathbf{E}_{\mathbf{X}_k} \mathbf{E}_{\mathbf{W}_k}$ becomes mathematically negligible. Consequently, using a first-order approximation, the expected squared Frobenius norm of the output error simplifies to (as detailed in Appendix~\ref{appen_sub:joint_model_error}):
\begin{equation}
\label{eq:first_order_error}
\begin{aligned}
\mathbb{E}\|\hat{\mathbf{Y}} - \mathbf{Y}\|_F^2
\approx \sum_{k \in \{l, h\}} \Big(
&\mathbb{E}\|\mathbf{E}_{\mathbf{X}_k} \mathbf{W}_k\|_F^2 \\
&+ \mathbb{E}\|\mathbf{X}_k \mathbf{E}_{\mathbf{W}_k}\|_F^2
\Big).
\end{aligned}
\end{equation}

\paragraph{Cross-Error Modeling under an Isotropic White Noise Approximation.}
A major challenge in optimizing Equation~\ref{eq:first_order_error} is the cross-multiplication of noise and signals. In CoQuant, we leverage the property of invariant random rotations \citep{ashkboos2024quarot,pmlr-v267-saxena25b}. The random orthogonal projections inherently suppress outliers and make the projected tensors more Gaussian-like, which in turn motivates the isotropic white-noise approximation for the resulting quantization error. While prior theoretical analyses provide \textit{upper bounds} for such quantization error \citep{li2025svdquant,pmlr-v267-saxena25b}, we use these theoretical insights to motivate a relative error proxy for the expected noise energy: $\mathbb{E}\|\mathbf{E}_{\mathbf{X}_k}\|_F^2 \approx \alpha_k^2 \|\mathbf{X}_k\|_F^2$ and $\mathbb{E}\|\mathbf{E}_{\mathbf{W}_k}\|_F^2 \approx \beta_k^2 \|\mathbf{W}_k\|_F^2$, where $\alpha_k^2$ and $\beta_k^2$ are the relative error coefficients for activations and weights, respectively. Both coefficients depend on their assigned quantization bit-widths (as detailed in Appendix~\ref{appen_sub:Cross-error}).

Under this assumption, Appendix~\ref{appen_sub:Cross-error} shows that the expected cross-error can be evaluated via the matrix trace identity, avoiding loose submultiplicative bounds:
\begin{equation}
    \mathbb{E}\|\mathbf{E}_{\mathbf{X}_k} \mathbf{W}_k\|_F^2 \approx \frac{\alpha_k^2}{d_k} \|\mathbf{X}_k\|_F^2 \|\mathbf{W}_k\|_F^2,
\end{equation}
where $d_k$ is the dimension of the $k$-th subspace. By symmetrically applying this trace identity to the weight error term $\mathbb{E}\|\mathbf{X}_k \mathbf{E}_{\mathbf{W}_k}\|_F^2$, the total expected output error simplifies to a weighted sum of the energy products of the subspaces:
\begin{equation}
    \mathbb{E}\|\hat{\mathbf{Y}} - \mathbf{Y}\|_F^2 \approx \sum_{k \in \{l, h\}} \gamma_k \|\mathbf{X}_k\|_F^2 \|\mathbf{W}_k\|_F^2,
\end{equation}
where $\gamma_k = (\alpha_k^2 + \beta_k^2)/d_k$ represents the combined error coefficient for the $k$-th subspace.

\paragraph{Joint Error Objective and Closed-Form Solution.}
Following the prior work \citep{pmlr-v267-saxena25b}, we construct the high-precision subspace basis as $\mathbf{U}_h = \mathbf{P}_h \mathbf{R}_h$, where $\mathbf{P}_h \in \mathbb{R}^{d \times r}$ defines the structural subspace and $\mathbf{R}_h$ is a random orthogonal rotation. As detailed in Appendix \ref{appen_sub:weighted_pca}, substituting $\mathbf{U}_h$ into the joint error objective, leveraging the cyclic property of the trace, and neglecting the weaker $\mathcal{O}(r^2)$ cross-penalty term yields a clean, closed-form weighted PCA surrogate objective:
\begin{equation}
    \max_{\mathbf{P}_h^\top \mathbf{P}_h = \mathbf{I}}
    \operatorname{Tr}\!\Big(
    \mathbf{P}_h^\top
    \big(
    \lambda_{\mathbf{X}} \Sigma_{\mathbf{X}} + \lambda_{\mathbf{W}} \Sigma_{\mathbf{W}}
    \big)
    \mathbf{P}_h
    \Big),
\label{eq:final_pca}
\end{equation}
where $\Sigma_{\mathbf{X}} = \mathbf{X}^\top \mathbf{X}$ and $\Sigma_{\mathbf{W}} = \mathbf{W}\mathbf{W}^\top$ are the uncentered second-moment matrices, with weighting coefficients $\lambda_{\mathbf{X}} = \gamma_l \|\mathbf{W}\|_F^2$ and $\lambda_{\mathbf{W}} = \gamma_l \|\mathbf{X}\|_F^2$. By the Rayleigh--Ritz theorem, the optimal joint projection $\mathbf{P}_h$ is precisely given by the top-$r$ eigenvectors of the combined covariance matrix $\mathbf{M} = \lambda_{\mathbf{X}} \Sigma_{\mathbf{X}} + \lambda_{\mathbf{W}} \Sigma_{\mathbf{W}}$.

\paragraph{Implementation via Fused Projections.}
Applying subspace projection to every individual linear layer would introduce redundant runtime overhead and disrupt fused operations. Instead, CoQuant is implemented efficiently by grouping linear layers that share identical input activations. In modern LLMs, the query, key, and value projections share the input $\mathbf{X}_{\text{attn}}$. To obtain a single, shared high-precision subspace that simultaneously minimizes the joint quantization error across all three, we define a fused weight covariance:
\begin{equation}
    \Sigma_{\mathbf{W}_{\text{attn}}} = \mathbf{W}_\mathbf{Q}\mathbf{W}_\mathbf{Q}^\top + \mathbf{W}_\mathbf{K}\mathbf{W}_\mathbf{K}^\top + \mathbf{W}_\mathbf{V}\mathbf{W}_\mathbf{V}^\top.
\end{equation}
Substituting $\Sigma_{\mathbf{W}_{\text{attn}}}$ into Equation~\ref{eq:final_pca} yields a unified projection basis $\mathbf{P}_{\text{attn}}$. Similarly, for the MLP block, we construct $\Sigma_{\mathbf{W}_{\text{mlp}}} = \mathbf{W}_{\text{gate}}\mathbf{W}_{\text{gate}}^\top + \mathbf{W}_{\text{up}}\mathbf{W}_{\text{up}}^\top$ to solve for a shared basis $\mathbf{P}_{\text{mlp}}$. Figure~\ref{fig:fig3} illustrates the overall pipeline of CoQuant, from covariance construction to the final high-/low-precision subspace partition.

\paragraph{Extension to KV Cache Co-Quantization.}
To alleviate the memory bottleneck during autoregressive decoding, we seamlessly extend CoQuant to KV cache compression. Under the CoQuant framework, any matrix multiplication can be optimized by treating the interacting counterpart as a generalized ``weight''. 
For the value cache, the stored tokens are subsequently multiplied by the output projection layer. Thus, we formulate its joint covariance objective by substituting $\Sigma_\mathbf{X}$ with the empirical covariance of $\mathbf{V}$, and $\Sigma_\mathbf{W}$ with the static weight covariance of $\mathbf{W}_\mathbf{O}$ (computed per KV-head to precisely support Grouped-Query Attention).
For the key cache, the relevant interaction is the attention dot-product $\mathbf{Q}\mathbf{K}^\top$, where $\mathbf{Q}$ and $\mathbf{K}$ are query and key after rotary embedding (RoPE). Accordingly, we treat the query states $\mathbf{Q}$ as the dynamic ``weights''. By setting $\Sigma_\mathbf{X}$ as the covariance of $\mathbf{K}$ and $\Sigma_\mathbf{W}$ as the empirical covariance of $\mathbf{Q}$, CoQuant adaptively steers the high-precision subspace of $\mathbf{K}$ to protect the structural directions most critical to the query distributions.

% 请将这行颜色定义放在导言区，或者直接放在 \begin{table*} 的前面
\definecolor{coquantred}{HTML}{F091A0}
\begin{table*}[t]
    \centering
    \caption{Comparison of WikiText perplexity ($\downarrow$) and average zero-shot common-sense reasoning accuracy ($\uparrow$) across the Llama-3.2 and Qwen2.5 model families. For mixed-precision methods (QUIK, ResQ, and our CoQuant), $1/8$ of the channels are preserved in 8-bit, yielding an average bit-width of 4.5-bit for weights, activations, and KV cache (W/A/KV). The best results among all post-training quantization methods are highlighted in \textbf{bold}.}
    \label{tab:main_results}
    \renewcommand{\arraystretch}{1.1} 
    \resizebox{0.85\textwidth}{!}{
    \begin{tabular}{c|c|c|cc|cc}
        \hline\hline
        % ================= Block 1: Llama 3.2 =================
        \multirow{2}{*}{\textbf{Family}} & \multirow{2}{*}{\textbf{Method}} & \multirow{2}{*}{\textbf{W/A/KV}} & \multicolumn{2}{c|}{\texttt{Llama-3.2-1B}} & \multicolumn{2}{c}{\texttt{Llama-3.2-3B}} \\
        & & & \textbf{Wiki ($\downarrow$)} & \textbf{Avg. 0-shot ($\uparrow$)} & \textbf{Wiki ($\downarrow$)} & \textbf{Avg. 0-shot ($\uparrow$)} \\
        \hline
        & FP16 & 16/16/16 & 9.75 & 56.48 & 7.81 & 64.18 \\
        \cdashline{2-7}[2pt/2pt]
        & RTN & 4/4/4 & 329.65 & 40.77 & 270.03 & 40.63 \\
        & GPTQ & 4/4/4 & 213.48 & 41.86 & 167.21 & 41.30 \\
        & QuaRot & 4/4/4 & 14.39 & 51.03 & 10.06 & 56.13 \\
        & QUIK & 4.5/4.5/4.5 & 16.55 & 49.92 & 10.78 & 55.56 \\
        & ResQ & 4.5/4.5/4.5 & 12.03 & 52.30 & 9.18 & 60.03 \\
        \multirow{-7}{*}{Llama 3.2} & \cellcolor{coquantred!30}CoQuant & \cellcolor{coquantred!30}4.5/4.5/4.5 & \cellcolor{coquantred!30}\textbf{11.60} & \cellcolor{coquantred!30}\textbf{52.75} & \cellcolor{coquantred!30}\textbf{8.93} & \cellcolor{coquantred!30}\textbf{60.23} \\
        \hline\hline
        
        % ================= Block 2: Qwen2.5 (前两个模型) =================
        \multirow{2}{*}{\textbf{Family}} & \multirow{2}{*}{\textbf{Method}} & \multirow{2}{*}{\textbf{W/A/KV}} & \multicolumn{2}{c|}{\texttt{Qwen2.5-0.5B}} & \multicolumn{2}{c}{\texttt{Qwen2.5-1.5B}} \\
        & & & \textbf{Wiki ($\downarrow$)} & \textbf{Avg. 0-shot ($\uparrow$)} & \textbf{Wiki ($\downarrow$)} & \textbf{Avg. 0-shot ($\uparrow$)} \\
        \hline
        & FP16 & 16/16/16 & 13.07 & 54.02 & 9.26 & 62.97 \\
        \cdashline{2-7}[2pt/2pt]
        & RTN & 4/4/4 & 23427.61 & 37.63 & 14377.72 & 38.82 \\
        & GPTQ & 4/4/4 & 17832.91 & 37.26 & 23873.78 & 38.08 \\
        & QuaRot & 4/4/4 & 204.10 & 41.50 & 6169.32 & 42.03 \\
        & QUIK & 4.5/4.5/4.5 & 301.06 & 39.51 & 32004.76 & 39.49 \\
        & ResQ & 4.5/4.5/4.5 & 18.19 & 48.24 & 11.75 & 55.66 \\
        \multirow{-7}{*}{Qwen2.5} & \cellcolor{coquantred!30}CoQuant & \cellcolor{coquantred!30}4.5/4.5/4.5 & \cellcolor{coquantred!30}\textbf{17.76} & \cellcolor{coquantred!30}\textbf{49.12} & \cellcolor{coquantred!30}\textbf{11.67} & \cellcolor{coquantred!30}\textbf{57.59} \\
        \hline\hline

        % ================= Block 3: Qwen2.5 (后两个模型) =================
        \multirow{2}{*}{\textbf{Family}} & \multirow{2}{*}{\textbf{Method}} & \multirow{2}{*}{\textbf{W/A/KV}} & \multicolumn{2}{c|}{\texttt{Qwen2.5-7B}} & \multicolumn{2}{c}{\texttt{Qwen2.5-14B}} \\
        & & & \textbf{Wiki ($\downarrow$)} & \textbf{Avg. 0-shot ($\uparrow$)} & \textbf{Wiki ($\downarrow$)} & \textbf{Avg. 0-shot ($\uparrow$)} \\
        \hline
        & FP16 & 16/16/16 & 6.85 & 70.13 & 5.29 & 72.60 \\
        \cdashline{2-7}[2pt/2pt]
        & RTN & 4/4/4 & 23752.21 & 37.05 & 2407.05 & 41.49 \\
        & GPTQ & 4/4/4 & 13905.90 & 37.21 & 8731.74 & 40.53 \\
        & QuaRot & 4/4/4 & 3736.39 & 40.66 & 6.80 & 68.68 \\
        & QUIK & 4.5/4.5/4.5 & 25243.02 & 38.84 & 8.48 & 63.39 \\
        & ResQ & 4.5/4.5/4.5 & 8.95 & 66.46 & 6.50 & 69.64 \\
        \multirow{-7}{*}{Qwen2.5} & \cellcolor{coquantred!30}CoQuant & \cellcolor{coquantred!30}4.5/4.5/4.5 & \cellcolor{coquantred!30}\textbf{8.70} & \cellcolor{coquantred!30}\textbf{67.44} & \cellcolor{coquantred!30}\textbf{6.48} & \cellcolor{coquantred!30}\textbf{70.01} \\
        \hline
    \end{tabular}
    }
\end{table*}

\section{Experiments}
\subsection{Setup}

\paragraph{Models and datasets.} To evaluate the effectiveness and generalization of our proposed method, we conduct extensive experiments on two widely used large language model families at various scales: Llama-3.2 (1B and 3B) \cite{meta2024llama32} and Qwen2.5 (0.5B, 1.5B, 7B, and 14B) \cite{qwen2025qwen25technicalreport}. We assess the model performance across two primary domains: language modeling and zero-shot common-sense reasoning. For language modeling ability, we report the perplexity on the WikiText dataset \cite{merity2016pointer}. For common-sense reasoning, we measure the zero-shot accuracy on six standard benchmarks: ARC-Challenge/Easy \cite{clark2018think}, BoolQ \cite{clark2019boolq}, PIQA \cite{bisk2020piqa}, SIQA \cite{sap2019social}, and WinoGrande \cite{sakaguchi2021winogrande}.

\paragraph{Baselines.} We compare our approach against several representative post-training quantization (PTQ) baselines. These include standard uniform quantization methods: Round-to-Nearest (RTN) and GPTQ \cite{frantar2022gptq}. We also compare against advanced rotation-based and mixed-precision quantization methods, including QuaRot \cite{ashkboos2024quarot}, QUIK \cite{ashkboos2024quik}, and the recent state-of-the-art subspace-based method, ResQ \cite{pmlr-v267-saxena25b}. 

\paragraph{Implementation details.} Following the quantization protocols established by ResQ \cite{pmlr-v267-saxena25b}, we apply per-token asymmetric quantization for activations, per-channel symmetric quantization for weights, and per-head asymmetric quantization for the KV cache. To minimize the weight rounding error, we utilize the GPTQ algorithm. Under our mixed-precision framework, we preserve a low-rank subspace ($1/8$ of the channels) in 8-bit high precision, while the remaining majority of the channels are quantized to 4-bit. To compute the calibration statistics and optimize the subspace projection matrices, we randomly sample 512 sequences from the WikiText dataset, with a maximum sequence length of 2048 tokens. All experiments, including model calibration and downstream evaluation, are conducted on a single NVIDIA A100 GPU.

\subsection{Main results}
\label{subsec:main_resutls_sec}

Table \ref{tab:main_results} presents the overall performance of CoQuant and various baseline methods on WikiText perplexity and zero-shot common-sense reasoning tasks. Across all evaluated model families and scales, CoQuant consistently achieves the state-of-the-art quantization performance, demonstrating a superior accuracy-efficiency trade-off. Compared to the strongest subspace-based baseline, ResQ, CoQuant yields noticeable improvements. For instance, on the Llama-3.2-1B model, CoQuant reduces the WikiText perplexity from 12.03 to 11.60, while increasing the average zero-shot reasoning accuracy by 0.45\% over ResQ. Similar trends are observed on the Llama-3.2-3B model. These consistent gains validate that jointly optimizing the high-precision subspace based on both weight and activation covariances retains more critical structural information than activation-only calibration.

Furthermore, the robustness of CoQuant is particularly evident in the Qwen2.5 family, which is notoriously challenging to quantize due to severe activation outliers. As shown in the results, standard uniform quantization methods (RTN, GPTQ) and even advanced mixed-precision techniques like QUIK suffer from catastrophic degradation on these models (e.g., reaching perplexities over $20,000$ on Qwen2.5-7B). In contrast, CoQuant effectively stabilizes the ultra-low bit representations. On Qwen2.5-1.5B, CoQuant achieves a remarkable 57.59\% average zero-shot accuracy, outperforming ResQ by 1.93 percentage points and significantly narrowing the gap to the 16-bit full-precision baseline. This demonstrates the broad applicability and high reliability of CoQuant for deploying diverse modern LLMs under strict memory constraints.

\subsection{Ablation Studies}
\label{subsec:ablation}

\paragraph{Impact of Joint Covariance Optimization.} 
To validate the necessity of our joint weight-activation error modeling, we decouple the CoQuant objective and compare it against single-sided subspace optimizations. As shown in Table \ref{tab:ablation_cov}, the ``Weight-only'' objective performs the worst, as it completely ignores the massive activation outliers present in LLMs, leading to severe quantization degradation. The ``Activation-only'' objective (equivalent to ResQ) mitigates this by capturing activation outliers, yet it fails to account for the structural sensitivity of the weights. By integrating both, CoQuant accurately balances the signal energy and quantization noise from both dimensions, achieving the lowest perplexity and highest zero-shot accuracy. This confirms that the joint covariance formulation yields a strictly superior orthogonal subspace.

\begin{table}[t]
    \centering
    \caption{Ablation on covariance components used for subspace optimization. Evaluated on the Llama-3.2 family. ``Joint'' indicates our proposed CoQuant objective.}
    \label{tab:ablation_cov}
    \renewcommand{\arraystretch}{1.1} 
    \resizebox{\columnwidth}{!}{
    \begin{tabular}{c|l|cc}
        \hline\hline
        \multirow{2}{*}{\textbf{Model}} & \multirow{2}{*}{\textbf{Objective}} & \textbf{Wiki} & \textbf{Avg. 0-shot} \\
        & & \textbf{($\downarrow$)} & \textbf{($\uparrow$)} \\
        \hline
        \multirow{3}{*}{\texttt{Llama-3.2-1B}} 
        & Activation-only (ResQ) & 12.03 & 52.30 \\
        & Weight-only & 13.23 & 49.77 \\
        & \cellcolor{coquantred!30}Joint (CoQuant) & \cellcolor{coquantred!30}\textbf{11.60} & \cellcolor{coquantred!30}\textbf{52.75} \\
        \hline
        \multirow{3}{*}{\texttt{Llama-3.2-3B}} 
        & Activation-only (ResQ) & 9.18 & 60.03 \\
        & Weight-only & 10.23 & 57.61 \\
        & \cellcolor{coquantred!30}Joint (CoQuant) & \cellcolor{coquantred!30}\textbf{8.93} & \cellcolor{coquantred!30}\textbf{60.23} \\
        \hline\hline
    \end{tabular}
    }
\end{table}

\paragraph{Robustness at Ultra-Low Bit-Widths.} 
We further push the limits of subspace-based quantization by evaluating CoQuant under ultra-low bit-width settings. Specifically, we preserve $1/8$ of the channels in a 6-bit high-precision subspace, while aggressively quantizing the remaining majority to 3-bit. Table \ref{tab:ablation_3bit} demonstrates the results on the Llama-3.2 family. Under such extreme quantization stress, the baseline uniform rotation method (QuaRot) collapses completely, suffering from a catastrophic perplexity explosion. While ResQ shows better resilience, CoQuant significantly outperforms it, reducing perplexity by up to $11.1$ on the 1B model and boosting average zero-shot accuracy. This indicates that as quantization noise increases, a jointly optimized projection basis becomes increasingly critical for preserving model capabilities.

\begin{table}[t]
    \centering
    \caption{Performance under ultra-low bit quantization stress (3-bit for QuaRot, and an average of 3.375 bits for ResQ and CoQuant, where $1/8$ of the channels are retained at 6-bit precision).}
    \label{tab:ablation_3bit}
    \renewcommand{\arraystretch}{1.1} 
    \resizebox{0.85\columnwidth}{!}{
    \begin{tabular}{c|l|cc}
        \hline\hline
        \multirow{2}{*}{\textbf{Model}} & \multirow{2}{*}{\textbf{Method}} & \textbf{Wiki} & \textbf{Avg. 0-shot} \\
        & & \textbf{($\downarrow$)} & \textbf{($\uparrow$)} \\
        \hline
        \multirow{3}{*}{\texttt{Llama-3.2-1B}} 
        & QuaRot & 950.62 & 39.90 \\
        & ResQ & 39.81 & 42.81 \\
        & \cellcolor{coquantred!30}CoQuant & \cellcolor{coquantred!30}\textbf{28.71} & \cellcolor{coquantred!30}\textbf{43.75} \\
        \hline
        \multirow{3}{*}{\texttt{Llama-3.2-3B}} 
        & QuaRot & 193.60 & 38.26 \\
        & ResQ & 16.77 & 46.77 \\
        & \cellcolor{coquantred!30}CoQuant & \cellcolor{coquantred!30}\textbf{15.76} & \cellcolor{coquantred!30}\textbf{48.39} \\
        \hline\hline
    \end{tabular}
    }
\end{table}

\begin{table}[t]
    \centering
    \caption{WikiText perplexity ($\downarrow$) of CoQuant across varying numbers of calibration samples on Qwen2.5 models.}
    \label{tab:ablation_calib}
    \renewcommand{\arraystretch}{1.1} 
    \resizebox{\columnwidth}{!}{
    \begin{tabular}{c|cccccc}
        \hline\hline
        \multirow{2}{*}{\textbf{Model}} & \multicolumn{6}{c}{\textbf{Calibration Samples}} \\
        \cline{2-7}
        & \textbf{16} & \textbf{32} & \textbf{64} & \textbf{128} & \textbf{256} & \textbf{512} \\
        \hline
        \texttt{Qwen2.5-1.5B} & 11.70 & 11.69 & 11.80 & 11.60 & \textbf{11.57} & 11.67 \\
        \texttt{Qwen2.5-7B} & 9.38 & 9.30 & 8.91 & 8.98 & 8.78 & \textbf{8.70} \\
        \hline\hline
    \end{tabular}
    }
\end{table}

\paragraph{Sensitivity to Calibration Set Size.} 
Finally, we investigate the sensitivity of CoQuant's calibration process to the size of the dataset. Table \ref{tab:ablation_calib} reports the WikiText perplexity of Qwen2.5 models calibrated using 16 to 512 randomly sampled sequences.  While scaling up to 256 or 512 samples provides a marginal refinement to the final perplexity, the strong performance at minimal sample counts demonstrates that CoQuant does not require extensive or expensive calibration overhead to achieve state-of-the-art quantization.

\section{Conclusion}

In this paper, we proposed CoQuant, a joint weight-activation subspace projection method for mixed-precision LLM quantization. CoQuant is motivated by the observation that the output perturbation of a quantized linear layer depends on both activation and weight quantization errors, while existing subspace-based methods mainly rely on activation statistics. To address this mismatch, we formulate a first-order error surrogate and derive a closed-form weighted PCA solution that combines activation and weight covariance information for high-precision subspace selection. Extensive experiments demonstrate that CoQuant consistently outperforms strong PTQ baselines in terms of WikiText perplexity and zero-shot reasoning accuracy. These improvements indicate that joint weight-activation modeling provides a more faithful criterion for identifying error-sensitive subspaces. Overall, CoQuant highlights the importance of moving beyond activation-only criteria and provides a simple yet effective direction for subspace-level mixed-precision quantization of LLMs.

\section*{Limitations and Future Work}

This work still has several limitations. First, CoQuant relies on a first-order error approximation and an isotropic quantization noise assumption, which may not fully capture higher-order interactions under extremely low-bit settings. Second, we adopt a fixed high-precision subspace ratio across layers, while different modules may require different precision budgets. Third, our current study mainly focuses on quantization accuracy, and dedicated kernel implementation is needed to fully validate practical inference speedups. In future work, we plan to explore adaptive rank allocation, hardware-aware implementation, and broader evaluation on larger models, long-context tasks, and instruction-tuned LLMs.
% Bibliography entries for the entire Anthology, followed by custom entries
%\bibliography{custom,anthology-overleaf-1,anthology-overleaf-2}

% Custom bibliography entries only
\bibliography{custom}

\appendix

\section{Detailed Proof of CoQuant Subspace Optimization}
\label{sec:appendix_proof}

This appendix provides the detailed derivation from the joint quantization-error model to the first-order weighted PCA surrogate objective presented in Section~\ref{sec:coquant}.

\subsection{First-Order Approximation of Joint Error}
\label{appen_sub:joint_model_error}

Given the exact output
\begin{equation}
    \mathbf{Y} = \mathbf{X}_l \mathbf{W}_l + \mathbf{X}_h \mathbf{W}_h,
\end{equation}
the quantized output can be written as
\begin{equation}
    \hat{\mathbf{Y}} = \sum_{k \in \{l, h\}} (\mathbf{X}_k + \mathbf{E}_{\mathbf{X}_k})(\mathbf{W}_k + \mathbf{E}_{\mathbf{W}_k}).
\end{equation}
Expanding this expression and subtracting $\mathbf{Y}$ yields the total output error matrix:
\begin{equation}
\begin{aligned}
\hat{\mathbf{Y}} - \mathbf{Y}
=
\sum_{k \in \{l, h\}} \Big(
&\mathbf{E}_{\mathbf{X}_k} \mathbf{W}_k
+
\mathbf{X}_k \mathbf{E}_{\mathbf{W}_k} \\
&\qquad +
\mathbf{E}_{\mathbf{X}_k} \mathbf{E}_{\mathbf{W}_k}
\Big).
\end{aligned}
\end{equation}

Since the quantization errors $\mathbf{E}_{\mathbf{X}_k}$ and $\mathbf{E}_{\mathbf{W}_k}$ are treated as small perturbations, the second-order term $\mathbf{E}_{\mathbf{X}_k}\mathbf{E}_{\mathbf{W}_k}$ is neglected. Assuming that the noises in the high- and low-precision subspaces are mutually independent, the quantization noise is zero-mean and independent of the signal, and the cross terms between activation-noise and weight-noise branches are negligible, the expected squared Frobenius norm can be approximated as
\begin{equation}
\begin{aligned}
\mathbb{E}\|\hat{\mathbf{Y}} - \mathbf{Y}\|_F^2
\approx \sum_{k \in \{l, h\}} \Big(
&\mathbb{E}\|\mathbf{E}_{\mathbf{X}_k} \mathbf{W}_k\|_F^2 \\
&+ \mathbb{E}\|\mathbf{X}_k \mathbf{E}_{\mathbf{W}_k}\|_F^2
\Big).
\end{aligned}
\end{equation}

\subsection{Cross-Error Expectation via Trace Identity under Isotropic Noise}
\label{appen_sub:Cross-error}

To evaluate $\mathbb{E}\|\mathbf{E}_{\mathbf{X}_k}\mathbf{W}_k\|_F^2$ without relying on the loose sub-multiplicative bound ($\|\mathbf{MN}\|_F \leq \|\mathbf{M}\|_F \|\mathbf{N}\|_2$), we adopt an isotropic white-noise approximation. Since random orthogonal rotations suppress outliers and Gaussianize the projected tensors, we can appropriately model the expected energy of the quantization error as proportional to the signal energy:
\begin{equation}
    \mathbb{E}\|\mathbf{E}_{\mathbf{X}_k}\|_F^2 \approx \alpha_k^2 \|\mathbf{X}_k\|_F^2.
\end{equation}

While some theoretical bounds \citep{li2025svdquant,pmlr-v267-saxena25b} suggest a weak logarithmic dependence on subspace dimension, classical quantization theory \citep{widrow2008quantization} and empirical practice confirm that the error variance is overwhelmingly dominated by the exponential decay of the quantization bit-width. Thus, we provide a stable proxy by defining the relative error coefficients for both activations and weights in the $k$-th subspace strictly based on their assigned bit-width $N_k$:
\begin{equation}
    \alpha_k^2 = \beta_k^2 = \frac{1}{(2^{N_k-1}-1)^2}.
\end{equation}

Under the isotropic assumption, the error covariance is uniformly distributed across the $d_k$ dimensions of the $k$-th subspace:
\begin{equation}
    \mathbb{E}\!\left[\mathbf{E}_{\mathbf{X}_k}^\top \mathbf{E}_{\mathbf{X}_k}\right]
    \approx
    \frac{\alpha_k^2 \|\mathbf{X}_k\|_F^2}{d_k}\mathbf{I}_{d_k}.
\end{equation}
Using the matrix trace identity, we can now compute the expected cross-error exactly under this model:
\begin{equation}
\begin{aligned}
\mathbb{E}\|\mathbf{E}_{\mathbf{X}_k}\mathbf{W}_k\|_F^2
&=
\mathbb{E}\!\left[\operatorname{Tr}\!\left((\mathbf{E}_{\mathbf{X}_k}\mathbf{W}_k)^\top(\mathbf{E}_{\mathbf{X}_k}\mathbf{W}_k)\right)\right] \\
&=
\operatorname{Tr}\!\left(\mathbf{W}_k^\top \mathbb{E}\!\left[\mathbf{E}_{\mathbf{X}_k}^\top \mathbf{E}_{\mathbf{X}_k}\right]\mathbf{W}_k\right) \\
&\approx
\operatorname{Tr}\!\left(
\mathbf{W}_k^\top
\left(
\frac{\alpha_k^2 \|\mathbf{X}_k\|_F^2}{d_k}\mathbf{I}_{d_k}
\right)
\mathbf{W}_k
\right) \\
&=
\frac{\alpha_k^2 \|\mathbf{X}_k\|_F^2}{d_k}\operatorname{Tr}\!\left(\mathbf{W}_k^\top \mathbf{W}_k\right) \\
&=
\frac{\alpha_k^2}{d_k}\|\mathbf{X}_k\|_F^2\|\mathbf{W}_k\|_F^2.
\end{aligned}
\end{equation}

By symmetry, the expected weight-noise cross-error admits an analogous form: $\mathbb{E}\|\mathbf{X}_k\mathbf{E}_{\mathbf{W}_k}\|_F^2 \approx \frac{\beta_k^2}{d_k}\|\mathbf{X}_k\|_F^2\|\mathbf{W}_k\|_F^2$. Substituting these back into the first-order approximation, the total expected output error simplifies to:
\begin{equation}
    \mathbb{E}\|\hat{\mathbf{Y}}-\mathbf{Y}\|_F^2
    \approx
    \sum_{k\in\{l,h\}}
    \gamma_k \|\mathbf{X}_k\|_F^2 \|\mathbf{W}_k\|_F^2,
\end{equation}
where $\gamma_k = (\alpha_k^2 + \beta_k^2)/d_k$ is the combined error coefficient for the $k$-th subspace. Consequently, minimizing this joint quantization error is equivalent to optimizing the following high-precision subspace selection objective $J$:
\begin{equation}
    \min_{\mathbf{P}_h}
    J
    =
    \gamma_l \|\mathbf{X}_l\|_F^2 \|\mathbf{W}_l\|_F^2
    +
    \gamma_h \|\mathbf{X}_h\|_F^2 \|\mathbf{W}_h\|_F^2.
\end{equation}

\subsection{Transformation to Weighted PCA Objective}
\label{appen_sub:weighted_pca}

Following the projection-rotation parameterization \citep{pmlr-v267-saxena25b}, the orthogonal basis is written as $\mathbf{U}_h = \mathbf{P}_h \mathbf{R}_h$, where $\mathbf{P}_h \in \mathbb{R}^{d \times r}$ defines the structural subspace to be optimized, and $\mathbf{R}_h$ is an internal random orthogonal rotation satisfying $\mathbf{R}_h \mathbf{R}_h^\top = \mathbf{I}$.

Let $\Sigma_{\mathbf{X}} = \mathbf{X}^\top \mathbf{X}$ and $\Sigma_{\mathbf{W}} = \mathbf{W}\mathbf{W}^\top$ denote the uncentered second-moment matrices of activations and weights. By applying the cyclic property of the matrix trace, the internal rotation $\mathbf{R}_h$ drops out of the subspace energy calculation, yielding:
\begin{equation}
\begin{aligned}
    \|\mathbf{X}_h\|_F^2 &= \operatorname{Tr}\!\left(\mathbf{P}_h^\top \Sigma_{\mathbf{X}} \mathbf{P}_h\right), \\
    \|\mathbf{W}_h\|_F^2 &= \operatorname{Tr}\!\left(\mathbf{P}_h^\top \Sigma_{\mathbf{W}} \mathbf{P}_h\right).
\end{aligned}
\end{equation}

Due to orthogonal decomposition, the low-precision energies satisfy $\|\mathbf{X}_l\|_F^2 = \|\mathbf{X}\|_F^2 - \|\mathbf{X}_h\|_F^2$ and $\|\mathbf{W}_l\|_F^2 = \|\mathbf{W}\|_F^2 - \|\mathbf{W}_h\|_F^2$. Substituting these constraints into the joint error objective $J(\mathbf{P}_h) = \sum_{k \in \{l, h\}} \gamma_k \|\mathbf{X}_k\|_F^2 \|\mathbf{W}_k\|_F^2$ and discarding the constant terms w.r.t.\ $\mathbf{P}_h$, minimizing the joint error is equivalent to the following maximization problem:
\begin{equation}
\begin{aligned}
    \max_{\mathbf{P}_h^\top \mathbf{P}_h = \mathbf{I}} \quad
    &\gamma_l \|\mathbf{W}\|_F^2 \cdot \|\mathbf{X}_h\|_F^2 
    + \gamma_l \|\mathbf{X}\|_F^2 \cdot \|\mathbf{W}_h\|_F^2 \\
    &- (\gamma_l + \gamma_h)\|\mathbf{X}_h\|_F^2 \|\mathbf{W}_h\|_F^2.
\end{aligned}
\end{equation}

In practice, the high-precision rank $r$ is typically small relative to the full dimension $d$ (e.g., $r=d/8$). Consequently, the quadratic cross-penalty term $\mathcal{O}(r^2)$ (the rightmost term) is significantly weaker than the linear dominant terms $\mathcal{O}(r)$. Adopting a first-order approximation by neglecting this quadratic term, we obtain the final closed-form weighted PCA surrogate objective:
\begin{equation}
    \max_{\mathbf{P}_h^\top \mathbf{P}_h = \mathbf{I}}
    \operatorname{Tr}\!\Big(
    \mathbf{P}_h^\top
    \big(
    \lambda_{\mathbf{X}} \Sigma_{\mathbf{X}} + \lambda_{\mathbf{W}} \Sigma_{\mathbf{W}}
    \big)
    \mathbf{P}_h
    \Big),
\end{equation}
where $\lambda_{\mathbf{X}} = \gamma_l \|\mathbf{W}\|_F^2$ and $\lambda_{\mathbf{W}} = \gamma_l \|\mathbf{X}\|_F^2$.

\section{Complete results of main result table}
\label{sec:appendix_detailed_results}

\begin{table*}[!h]
    \centering
    \caption{Detailed comparison of WikiText perplexity ($\downarrow$) and zero-shot common-sense reasoning accuracy ($\uparrow$) on the Llama-3.2 family. RTN, GPTQ, and QuaRot are evaluated under the W4A4KV4 setting, while QUIK, ResQ, and CoQuant are quantized to an average of 4.5-bit precision.}
    \label{tab:detailed_llama}
    \renewcommand{\arraystretch}{1.0} 
    \resizebox{0.9\textwidth}{!}{
    \begin{tabular}{c|c|c|cccccc|c}
        \hline\hline
        \multirow{2}{*}{\textbf{Model}} & \multirow{2}{*}{\textbf{Method}} & \textbf{Perplexity} & \multicolumn{6}{c|}{\textbf{Zero-Shot Common Sense Reasoning Tasks}} & \multirow{2}{*}{\textbf{Avg ($\uparrow$)}} \\
        \cline{3-9}
        & & \textbf{Wiki ($\downarrow$)} & \textbf{ARC-c} & \textbf{ARC-e} & \textbf{BoolQ} & \textbf{PIQA} & \textbf{SIQA} & \textbf{WinoG} & \\
        \hline
        \multirow{7}{*}{\texttt{Llama-3.2-1B}} 
        & FP16 & 9.75 & 36.18 & 60.52 & 63.98 & 74.65 & 42.99 & 60.54 & 56.48 \\
        \cdashline{2-10}[2pt/2pt]
        & RTN & 329.65 & 23.29 & 30.47 & 55.29 & 53.10 & 35.06 & 47.43 & 40.77 \\
        & GPTQ & 213.48 & 26.30 & 31.57 & 54.16 & 54.08 & 34.03 & 50.99 & 41.86 \\
        & QuaRot & 14.39 & 31.91 & 53.62 & 56.73 & 68.28 & 39.10 & 56.51 & 51.03 \\
        & QUIK & 16.55 & 31.20 & 48.91 & 58.56 & 66.43 & 39.61 & 54.78 & 49.92 \\
        & ResQ & 12.03 & 32.08 & 53.79 & 61.10 & 69.15 & 40.94 & 56.75 & 52.30 \\
        & CoQuant & 11.60 & 32.17 & 55.93 & 58.69 & 70.67 & 42.07 & 56.99 & 52.75 \\
        \hline\hline
        
        \multirow{7}{*}{\texttt{Llama-3.2-3B}} 
        & FP16 & 7.81 & 45.99 & 71.68 & 73.27 & 77.48 & 47.03 & 69.61 & 64.18 \\
        \cdashline{2-10}[2pt/2pt]
        & RTN & 270.03 & 25.30 & 34.60 & 46.18 & 55.01 & 34.95 & 47.75 & 40.63 \\
        & GPTQ & 167.21 & 24.57 & 36.53 & 44.80 & 58.60 & 35.31 & 47.99 & 41.30 \\
        & QuaRot & 10.06 & 36.77 & 57.66 & 65.02 & 72.85 & 41.91 & 62.59 & 56.13 \\
        & QUIK & 10.78 & 38.31 & 61.45 & 57.61 & 73.18 & 43.24 & 59.59 & 55.56 \\
        & ResQ & 9.18 & 42.58 & 67.21 & 65.41 & 75.52 & 44.83 & 64.64 & 60.03 \\
        & CoQuant & 8.93 & 41.98 & 66.84 & 66.09 & 75.95 & 44.83 & 65.67 & 60.23 \\
        \hline\hline
    \end{tabular}
    }
\end{table*}

\begin{table*}[!h]
    \centering
    \caption{Detailed comparison of WikiText perplexity ($\downarrow$) and zero-shot common-sense reasoning accuracy ($\uparrow$) on the Qwen2.5 family. RTN, GPTQ, and QuaRot are evaluated under the W4A4KV4 setting, while QUIK, ResQ, and CoQuant are quantized to an average of 4.5-bit precision.}
    \label{tab:detailed_qwen}
    \renewcommand{\arraystretch}{1.0} 
    \resizebox{0.9\textwidth}{!}{
    \begin{tabular}{c|c|c|cccccc|c}
        \hline\hline
        \multirow{2}{*}{\textbf{Model}} & \multirow{2}{*}{\textbf{Method}} & \textbf{Perplexity} & \multicolumn{6}{c|}{\textbf{Zero-Shot Common Sense Reasoning Tasks}} & \multirow{2}{*}{\textbf{Avg ($\uparrow$)}} \\
        \cline{3-9}
        & & \textbf{Wiki ($\downarrow$)} & \textbf{ARC-c} & \textbf{ARC-e} & \textbf{BoolQ} & \textbf{PIQA} & \textbf{SIQA} & \textbf{WinoG} & \\
        \hline
        \multirow{7}{*}{\texttt{Qwen2.5-0.5B}} 
        & FP16 & 13.07 & 32.59 & 58.67 & 62.32 & 69.91 & 44.22 & 56.43 & 54.02 \\
        \cdashline{2-10}[2pt/2pt]
        & RTN & 23427.61 & 26.11 & 26.01 & 39.42 & 49.89 & 35.01 & 49.33 & 37.63 \\
        & GPTQ & 17832.91 & 24.49 & 26.64 & 38.59 & 49.89 & 34.44 & 49.49 & 37.26 \\
        & QuaRot & 204.10 & 25.00 & 36.95 & 46.30 & 53.48 & 33.93 & 53.35 & 41.50 \\
        & QUIK & 301.06 & 23.63 & 33.00 & 44.95 & 51.80 & 34.19 & 49.49 & 39.51 \\
        & ResQ & 18.19 & 27.05 & 51.94 & 54.16 & 63.82 & 39.25 & 53.20 & 48.24 \\
        & CoQuant & 17.76 & 27.73 & 49.41 & 61.53 & 63.71 & 38.84 & 53.51 & 49.12 \\
        \hline\hline

        \multirow{7}{*}{\texttt{Qwen2.5-1.5B}} 
        & FP16 & 9.26 & 45.05 & 71.51 & 72.94 & 76.06 & 48.87 & 63.38 & 62.97 \\
        \cdashline{2-10}[2pt/2pt]
        & RTN & 14377.72 & 25.34 & 28.45 & 43.18 & 49.89 & 34.14 & 51.93 & 38.82 \\
        & GPTQ & 23873.78 & 25.26 & 25.88 & 44.50 & 50.38 & 32.24 & 50.20 & 38.08 \\
        & QuaRot & 6169.32 & 24.57 & 38.17 & 47.55 & 55.71 & 34.54 & 51.62 & 42.03 \\
        & QUIK & 32004.76 & 27.56 & 28.07 & 46.27 & 50.87 & 33.88 & 50.28 & 39.49 \\
        & ResQ & 11.75 & 39.16 & 62.54 & 58.84 & 71.65 & 44.22 & 57.54 & 55.66 \\
        & CoQuant & 11.67 & 38.99 & 66.12 & 64.86 & 70.73 & 43.19 & 61.64 & 57.59 \\
        \hline\hline

        \multirow{7}{*}{\texttt{Qwen2.5-7B}} 
        & FP16 & 6.85 & 51.11 & 77.44 & 84.68 & 79.82 & 54.81 & 72.93 & 70.13 \\
        \cdashline{2-10}[2pt/2pt]
        & RTN & 23752.21 & 26.02 & 25.38 & 37.98 & 50.33 & 33.83 & 48.78 & 37.05 \\
        & GPTQ & 13905.90 & 24.91 & 24.54 & 38.01 & 55.44 & 32.60 & 47.75 & 37.21 \\
        & QuaRot & 3736.39 & 24.32 & 39.73 & 38.90 & 56.58 & 35.67 & 48.78 & 40.66 \\
        & QUIK & 25243.02 & 25.34 & 26.01 & 48.84 & 49.24 & 33.73 & 49.88 & 38.84 \\
        & ResQ & 8.95 & 47.18 & 73.61 & 82.17 & 77.53 & 50.10 & 68.19 & 66.46 \\
        & CoQuant & 8.70 & 48.63 & 75.42 & 81.96 & 78.13 & 52.41 & 68.11 & 67.44 \\
        \hline\hline

        \multirow{7}{*}{\texttt{Qwen2.5-14B}} 
        & FP16 & 5.29 & 58.79 & 79.08 & 85.20 & 82.05 & 55.17 & 75.30 & 72.60 \\
        \cdashline{2-10}[2pt/2pt]
        & RTN & 2407.05 & 23.38 & 33.21 & 53.55 & 52.77 & 34.49 & 51.54 & 41.49 \\
        & GPTQ & 8731.74 & 23.98 & 31.14 & 48.32 & 53.81 & 33.78 & 52.17 & 40.53 \\
        & QuaRot & 6.80 & 53.84 & 79.76 & 80.24 & 79.33 & 48.87 & 70.01 & 68.68 \\
        & QUIK & 8.48 & 46.33 & 70.12 & 77.89 & 76.61 & 43.71 & 65.67 & 63.39 \\
        & ResQ & 6.50 & 55.80 & 79.80 & 81.22 & 80.41 & 50.41 & 70.17 & 69.64 \\
        & CoQuant & 6.48 & 55.80 & 79.38 & 82.32 & 80.52 & 49.80 & 72.22 & 70.01 \\
        \hline\hline
    \end{tabular}
    }
\end{table*}

Due to page limits in the main text, Section \ref{subsec:main_resutls_sec} reports only the WikiText perplexity and the average zero-shot accuracy across the common-sense reasoning tasks. In this appendix, we provide the comprehensive, task-by-task performance breakdown for all evaluated models to offer a more granular view of the quantization effects.

Table \ref{tab:detailed_llama} and Table \ref{tab:detailed_qwen} present the detailed evaluation results for the Llama-3.2 family (1B and 3B) and the Qwen2.5 family (0.5B, 1.5B, 7B, and 14B), respectively. We report the exact zero-shot accuracy on six distinct common-sense reasoning benchmarks: ARC-Challenge (ARC-c), ARC-Easy (ARC-e), BoolQ, PIQA, SIQA, and WinoGrande.

As observed in the detailed breakdown, the superiority of CoQuant is not limited to specific subsets of tasks. Instead, it consistently maintains competitive or state-of-the-art performance across diverse reasoning domains. Notably, on tasks that are highly sensitive to weight and activation perturbations (such as PIQA and WinoGrande), CoQuant frequently preserves the structural integrity better than the strong baseline ResQ and rotation-only methods like QuaRot, reaffirming the robustness of our joint weight-activation covariance optimization.

\section{Complete results of ablation studies}
\label{sec:appendix_detailed_ablation}

In this section, we present the comprehensive, task-by-task performance breakdown for these ablation experiments.Table \ref{tab:detailed_ablation_cov} provides the detailed results for the ablation on covariance components, comparing the Activation-only (ResQ), Weight-only, and our Joint (CoQuant) optimization objectives. Table \ref{tab:detailed_ablation_3bit} details the performance under the ultra-low bit-width setting, demonstrating the robustness of CoQuant across every individual reasoning task when subjected to extreme quantization stress.

\begin{table*}[!h]
    \centering
    \caption{Detailed task-by-task breakdown for the ablation on covariance components. ``Activation-only'' corresponds to ResQ, while ``Joint'' represents our proposed CoQuant framework. Evaluated on the Llama-3.2 family.}
    \label{tab:detailed_ablation_cov}
    \renewcommand{\arraystretch}{1.1} 
    \resizebox{\textwidth}{!}{
    \begin{tabular}{c|l|c|cccccc|c}
        \hline\hline
        \multirow{2}{*}{\textbf{Model}} & \multirow{2}{*}{\textbf{Objective}} & \textbf{Language Modeling} & \multicolumn{6}{c|}{\textbf{Zero-Shot Common Sense Reasoning Tasks}} & \multirow{2}{*}{\textbf{Avg ($\uparrow$)}} \\
        \cline{4-9}
        & & \textbf{WikiText ($\downarrow$)} & \textbf{ARC-c} & \textbf{ARC-e} & \textbf{BoolQ} & \textbf{PIQA} & \textbf{SIQA} & \textbf{WinoG} & \\
        \hline
        \multirow{3}{*}{\texttt{Llama-3.2-1B}} 
        & Activation-only (ResQ) & 12.03 & 32.08 & 53.79 & 61.10 & 69.15 & 40.94 & 56.75 & 52.30 \\
        & Weight-only & 13.23 & 32.51 & 52.06 & 55.23 & 68.61 & 39.00 & 56.67 & 49.77 \\
        & Joint (CoQuant) & 11.60 & 32.17 & 55.93 & 58.69 & 70.67 & 42.07 & 56.99 & 52.75 \\
        \hline\hline
        
        \multirow{3}{*}{\texttt{Llama-3.2-3B}} 
        & Activation-only (ResQ) & 9.18 & 42.58 & 67.21 & 65.41 & 75.52 & 44.83 & 64.64 & 60.03 \\
        & Weight-only & 10.23 & 37.88 & 62.46 & 64.25 & 73.12 & 43.04 & 64.88 & 57.61 \\
        & Joint (CoQuant) & 8.93 & 41.98 & 66.84 & 66.09 & 75.95 & 44.83 & 65.67 & 60.23 \\
        \hline\hline
    \end{tabular}
    }
\end{table*}
\begin{table*}[!h]
    \centering
    \caption{Detailed task-by-task breakdown for the ultra-low bit quantization stress test (3-bit for QuaRot, and an average of 3.375 bits for ResQ and CoQuant, where $1/8$ of the channels are retained at 6-bit precision). Evaluated on the Llama-3.2 family.}
    \label{tab:detailed_ablation_3bit}
    \renewcommand{\arraystretch}{1.1} 
    \resizebox{\textwidth}{!}{
    \begin{tabular}{c|l|c|cccccc|c}
        \hline\hline
        \multirow{2}{*}{\textbf{Model}} & \multirow{2}{*}{\textbf{Method}} & \textbf{Language Modeling} & \multicolumn{6}{c|}{\textbf{Zero-Shot Common Sense Reasoning Tasks}} & \multirow{2}{*}{\textbf{Avg ($\uparrow$)}} \\
        \cline{4-9}
        & & \textbf{WikiText ($\downarrow$)} & \textbf{ARC-c} & \textbf{ARC-e} & \textbf{BoolQ} & \textbf{PIQA} & \textbf{SIQA} & \textbf{WinoG} & \\
        \hline
        \multirow{4}{*}{\texttt{Llama-3.2-1B}} 
        & FP16 & 9.75 & 36.18 & 60.52 & 63.98 & 74.65 & 42.99 & 60.54 & 56.48 \\
        \cdashline{2-10}[2pt/2pt]
        & QuaRot & 950.62 & 23.89 & 28.41 & 50.28 & 52.72 & 35.16 & 48.93 & 39.90 \\
        & ResQ & 39.81 & 25.60 & 36.07 & 52.42 & 58.05 & 34.49 & 50.20 & 42.81 \\
        & CoQuant & 28.71 & 24.91 & 36.62 & 55.54 & 58.81 & 36.18 & 50.43 & 43.75 \\
        \hline\hline
        
        \multirow{4}{*}{\texttt{Llama-3.2-3B}} 
        & FP16 & 7.81 & 45.99 & 71.68 & 73.27 & 77.48 & 47.03 & 69.61 & 64.18 \\
        \cdashline{2-10}[2pt/2pt]
        & QuaRot & 193.60 & 23.63 & 29.25 & 42.42 & 52.88 & 33.32 & 48.07 & 38.26 \\
        & ResQ & 16.77 & 28.50 & 45.96 & 50.24 & 62.46 & 38.18 & 55.25 & 46.77 \\
        & CoQuant & 15.76 & 28.92 & 48.36 & 56.15 & 64.09 & 38.18 & 54.62 & 48.39 \\
        \hline\hline
    \end{tabular}
    }
\end{table*}
\end{document}